# Design, Results and Industry Implications of the World's First Insurance Large Language Model Evaluation Benchmark

Hua Zhou, Central University of Finance and Economics

Bing Ma, Central University of Finance and Economics

Yufei Zhang, Zetavision AI Lab

Yi Zhao, Zetavision AI Lab

## Abstract

The in-depth application of large language models (LLMs) in vertical fields is facing severe challenges regarding professionalism and reliability. Due to the threefold characteristics of the insurance industry—"strong regulatory constraints, high professional barriers, and high-risk sensitivity"—the evaluation of models in this sector has long lacked a systematic, reproducible, and authoritative framework for model evaluation. To address this gap, the School of Insurance and the China Institute of Actuarial Science at Central University of Finance and Economics, leveraging over seventy years of profound accumulation in insurance and actuarial disciplines, have designed and open-sourced the world's first professional evaluation benchmark for large models deeply integrated with the insurance knowledge system: the CUFE Insurance Evaluation Suite (CUFEInse). The data and code are available at https://github.com/CUFEInse/CUFEInse.

This paper comprehensively elaborates on the construction methodology, multi-dimensional evaluation system, and underlying design philosophy of CUFEInse v1.0. Adhering to the principles of "quantitative-oriented, expert-driven, and multi-validation," the benchmark establishes an evaluation framework covering 5 core dimensions, 54 sub-indicators, and 14,430 high-quality questions, encompassing insurance theoretical knowledge, industry understanding, safety and compliance, intelligent agent application, and logical rigor. Based on this benchmark, a comprehensive evaluation was conducted on 11 mainstream large language models.

The evaluation results reveal that general-purpose models suffer from common bottlenecks such as weak actuarial capabilities and inadequate compliance adaptation. High-quality domain-specific training demonstrates significant advantages in insurance vertical scenarios but exhibits shortcomings in business adaptation and compliance. The evaluation also accurately identifies the common bottlenecks of current large models in professional scenarios such as insurance actuarial, underwriting and claim settlement reasoning, and compliant marketing copywriting. The establishment of CUFEInse not only fills the gap in professional evaluation benchmarks for the insurance field, providing academia and industry with a professional, systematic, and authoritative evaluation tool, but also its construction concept and methodology offer important references for the evaluation paradigm of large models in vertical fields, serving as an authoritative reference for academic model optimization and industrial model selection. Finally, the paper looks forward to the future iteration direction of the evaluation benchmark and the core development direction of "domain adaptation + reasoning enhancement" for insurance large models.

**Keywords:** Insurance Large Language Model; Evaluation Benchmark; CUFEInse; Actuarial Evaluation; Compliance and Safety; Intelligent Agent Application; Domain Adaptation; Central University of Finance and Economics; School of Insurance

# 1 Introduction

## 1.1 Research Background: Adaptability Challenges Between Large Models and the Insurance Industry

In recent years, the rapid development of large language model technology has been reshaping various industries with unprecedented depth and breadth. In highly specialized vertical fields such as finance, healthcare, and law, how to objectively and accurately evaluate the professional capabilities of large models has become a key prerequisite for promoting their safe and reliable implementation. As one of the three pillars of the modern financial system, the insurance industry features a long business chain, complex professional knowledge systems, strict regulatory requirements, and direct relevance to national economy, people's livelihood, and social stability. Its particularities are reflected in three aspects:

1) **Professional Knowledge Intensiveness:** It covers interdisciplinary knowledge such as insurance science, actuarial mathematics, insurance law, and social insurance policies. Moreover, actuarial calculations (e.g., reserve assessment, premium pricing) require rigorous mathematical logic and industry experience;

2) **Rigid Regulatory Constraints:** From the compliance of marketing copy to the underwriting and claim settlement processes, all aspects must comply with policy requirements such as the “*Insurance Law*” and “*Administrative Measures for Personal Insurance Sales Behaviors*”. Non-compliance risks are directly linked to enterprise compliance costs and industry credibility;

3) **Complexity of Business Scenarios:** It involves the entire chain of "insured group positioning - insurance liability analysis - underwriting decision - claim settlement - post-insurance service," requiring models to possess scenario-based understanding and decision-making capabilities (e.g., risk prediction based on medical documents, insurance product comparison and recommendation).

These characteristics make general-purpose large models face a series of challenges such as insufficient professional knowledge, lack of logical rigor, and high compliance risks when directly applied to insurance scenarios. Although existing financial evaluation benchmarks [14,15] provide references for vertical fields, they have significant limitations[1,2]: first, fragmented knowledge coverage, focusing more on general financial scenarios such as banking and securities, without systematically integrating core areas such as insurance actuarial and compliance safety; second, single evaluation dimension, lacking in-depth investigation and special consideration of core insurance business logic (e.g., actuarial, underwriting and claim settlement); third, insufficient expert participation, with question design failing to fully align with the actual needs of insurance business. To sum up, a professional evaluation benchmark that fully covers the insurance theoretical knowledge system, emphasizes professional rigor, and is expert-driven is still lacking.

## 1.2 Research Significance: Academic and Industrial Value of the CUFEInse Benchmark

To fill this gap, the School of Insurance and the China Institute of Actuarial Science at Central University of Finance and Economics, based on their profound disciplinary accumulation, jointly designed and open-sourced the "CUFEInse" insurance domain evaluation benchmark. As the world's first professional evaluation benchmark for large models deeply integrated with the insurance discipline system and fully covering insurance professional theoretical knowledge and application scenarios, it pioneers the systematic evaluation of large models in the insurance field.

The establishment of CUFEInse aims to provide academia and industry with a more professional, systematic, and authoritative evaluation tool for insurance large models, facilitating model research and development, industry selection, and application standardization, and ultimately promoting the digital transformation and intelligent upgrading of the insurance industry. Its core values are reflected in three aspects:

1) **Academic Aspect:** For the first time, it constructs a "insurance discipline knowledge - business scenario - technical capability" trinity evaluation framework, providing a standardized tool for the training optimization and technical route comparison of insurance large models;

2) **Industrial Aspect:** By quantifying the model's capabilities in key scenarios such as actuarial calculation, compliant copywriting generation, and underwriting and claim settlement decision-making, it provides data support for insurance companies' model selection and technology enterprises' product iteration;

3) **Regulatory Aspect:** Embedding insurance compliance requirements into the evaluation dimensions helps regulatory authorities grasp the application risks of LLMs in the insurance field and promote the coordinated development of "safety compliance + technological innovation" in the industry.

# 2 Design of the CUFEInse Insurance Large Model Evaluation Benchmark

## 2.1 Core Design Concept and Methodology

CUFEInse follows the core methodology of "quantitative-oriented, expert-driven, and multi-validation," ensuring the professionalism and credibility of the evaluation content through a four-stage process:

**Quantitative-oriented:** The benchmark adheres to objective and quantifiable questions as the main carrier of evaluation, avoiding biases caused by subjective evaluation and ensuring the reproducibility and comparability of evaluation results.

**Expert-driven:** The entire benchmark construction process is led by the full-time faculty team of the School of Insurance and the China Institute of Actuarial Science at Central University of Finance and Economics. These experts not only possess profound theoretical literacy but also most have rich experience in industry consulting and practice, ensuring the professionalism of the question content and its alignment with business scenarios.

**Multi-validation:** To ensure the quality of each question, the benchmark construction undergoes a strict four-step process:

1) **Industry Research (Demand Anchoring):** Joint research was conducted with leading insurance companies and insurance technology enterprises to identify 26 high-frequency business pain points such as "actuarial pricing," "claim dispute handling," and "compliant copywriting generation," and clarify the priority of evaluation dimensions;

2) **Expert Question Setting (Content Generation):** The expert team set questions based on the standardized insurance and actuarial discipline system, ensuring the systematicness and accuracy of knowledge points. The questions cover core knowledge modules of the insurance discipline (insurance products, insurance law, actuarial mathematics, etc.);

3) **Cross-review (Quality Control):** A "double-blind review" mechanism was adopted, where question-setting experts and industry practice experts independently reviewed the questions to eliminate ambiguous and off-syllabus questions;

4) **Sensitivity Review (Risk Avoidance):** Regarding sensitive content such as insurance compliance and customer privacy, the regulatory policy research team was invited to review whether the questions comply with the latest regulatory requirements, avoiding compliance risks caused by evaluation data.

## 2.2 Multi-dimensional Evaluation System Architecture

CUFEInse v1.0 includes 14,430 high-quality questions, covering question types such as single-choice, multiple-choice, judgment, short answer, reasoning and planning, retrieval and question answering, and label extraction. It fully covers the three core needs of models: "knowledge reserve - reasoning ability - scenario adaptation," and can comprehensively evaluate the model's knowledge reserve, reasoning ability, and scenario adaptability.

The five first-level dimensions and second-level subcategories are designed as follows:

**Table 1 Dimension and Subcategory Distribution of CUFEInse v1.0 Evaluation Set**

| First-Level Dimension | Core Second-Level Subcategories | Number of Questions | Proportion | Evaluation Objective |
|---|---|---|---|---|
| **Insurance Theoretical Knowledge** | Insurance Studies, Insurance Products, Insurance Law, Insurance Actuarial Science, Insurance Finance & Investment, Insurance Market & Operations, Social Insurance & Policy Insurance, etc. (14 subcategories) | 5,254 | 36.43 % | Evaluate the model's foundational cognition and professional depth of the insurance discipline system. |
| **Insurance Industry Understanding** | Target Customer Profiling, Insurance Liability Analysis, Underwriting & Claims Process, Policy Application, Underwriting, Claims Settlement, Payment, Post-Sale Operations, etc. (18 subcategories) | 4,448 | 30.84 % | Evaluate the model's scenario-specific understanding of the entire insurance business chain. |
| **Insurance Safety & Compliance** | Insurance Values, Insurance Technology & Ethics, Copy Compliance, Safety Bottom Line, etc. (5 subcategories) | 1,331 | 9.23 % | Evaluate the model's adherence to regulatory policies and professional ethics, avoiding compliance risks. |
| **Insurance Agent Application** | Product Comparison, Product Recommendation, Underwriting Decision Reasoning, Claims Decision Reasoning, Planning & Configuration, Insurance Tool Invocation, etc. (12 subcategories) | 2,081 | 14.43 % | Evaluate the model's ability as an "Insurance Agent" for intent understanding, decision suggestion, and tool coordination. |
| **Insurance Logical Rigor** | Insurance Problem Identification, Benefit Calculation, Clause Interpretation, Financial Numerical Calculation, etc. (5 subcategories) | 1,316 | 9.13 % | Evaluate the model's ability for logical decomposition and precise interpretation of complex actuarial calculations and clause details. |

## 2.3 Scoring Mechanism: Fairness Design of "Equal Weight for Dimensions + Balance for Subcategories"

To ensure the comprehensiveness, fairness, and interpretability of the evaluation, CUFEInse adopts a comprehensive scoring strategy of "equal weight for dimensions and balance for subcategories."

**Equal Weight for Dimensions:** The five first-level dimensions (theoretical knowledge, industry understanding, safety and compliance, intelligent agent application, and rigor) have equal weights in the comprehensive score. This design avoids the overall score being dominated by a single dimension (such as mere knowledge memory) and forces the model to achieve balanced development in multiple aspects such as knowledge, application, compliance, and logic to obtain high scores.

**Balance for Subcategories:** Under each first-level dimension, the questions of the second-level subcategories are evenly distributed according to their knowledge granularity. This means that the model's capabilities in each segmented knowledge field can be fairly reflected in the final score, preventing the possibility of "score inflation" due to excessive questions in certain subcategories. This design ensures that the evaluation results have good interpretability and comparability, and can clearly reflect the differences and shortcomings of the model's capabilities in specific knowledge fields.

Finally, the scores of all models are standardized using a 100-point scale for intuitive comparison.

## 2.4 Key Technological Innovations of the Benchmark

Compared with existing evaluation benchmarks in the financial field such as C-Eval[11] and AGIEval[12], CUFEInse has achieved key technological innovations in five aspects:

1) **Systematic Sorting and Question Setting of the Insurance Discipline Knowledge System:**

We organized full-time teachers from the School of Insurance and the China Institute of Actuarial Science at Central University of Finance and Economics to systematically classify, sort, and set questions on insurance theoretical knowledge based on the standardized insurance and actuarial discipline system (covering insurance products, insurance systems and principles, insurance science, insurance market and operation, insurance law, insurance actuarial, insurance finance and investment, social insurance and policy insurance). This ensures high standards in the disciplinary professionalism and knowledge system integrity of the evaluation content, avoiding the problems of fragmented and insufficiently in-depth insurance professional knowledge in previous evaluations.

2) **Multi-level Classification System and Question Proportion Optimization:**

We revised and carefully adjusted the first-level and second-level classifications one by one, and optimized the proportion distribution of questions in different categories to better align with the knowledge distribution and importance ratio in actual insurance business scenarios. This fine-grained classification and weight design ensure that the evaluation results can more truly reflect the differences in the model's strengths and weaknesses in various segmented capabilities in the insurance field.

3) **Emphasis on Logical Rigor and Safety Compliance:**

Special dimensions of "Logical Rigor" and "Compliance and Safety" are added, focusing on the model's reasoning chain of thought [13], output stability, authenticity (avoiding hallucinations) in insurance scenarios, and its ability to abide by insurance regulatory policies, professional ethics, and ethical bottom lines. This is highly consistent with the characteristics of strict regulation and risk sensitivity in the insurance industry.

4) **Inclusion of Insurance Actuarial and Professional Certification Examination Content:**

Ensure the accuracy and professionalism of actuarial mathematical formulas, symbol expressions, and calculation processes, solving the problems of format confusion and semantic understanding deviation that large models may encounter when processing complex actuarial calculations. At the same time, the simulation of high-standard and standardized examinations effectively evaluates whether the model has the professional knowledge level equivalent to insurance professionals. This technological innovation improves the evaluation accuracy of the model's ability to process professional mathematical and financial content.

5) **Model Architecture Adaptability and Computational Efficiency Optimization:**

We designed a multi-scale evaluation scheme based on the characteristics of different model architectures, which can not only comprehensively evaluate the professional capabilities of super-large parameter models but also fully consider the deployment needs of lightweight models in actual insurance business scenarios. The evaluation framework supports computational efficiency evaluation, focusing on the response speed and resource consumption of the model in typical insurance business scenarios, providing practical references for industrial model selection.

# 3 Implementation and Multi-dimensional In-depth Analysis of CUFEInse v1.0 Evaluation

## 3.1 Evaluation Methodology: Scientific Guarantee of "Model Selection + Process Control"

### 3.1.1 Scope of Evaluated Models

In this evaluation, 11 representative large models were selected, classified by "model type - open-source attribute," covering the main directions of current large model technology routes:

- General closed-source models: Gemini-2.5-Pro-0617[3]、GPT-4o-1120[4]
- General open-source models: DeepSeek-R1-0528[5]、GPT-oss-120b[6]、Qwen3-235B-A22B-instruct[7] 、 Qwen3-235B-A22B-think[7] 、 Qwen3-32B-instruct[7] 、 Qwen3-32B-think[7]
- Domain-specific models: AntGroup Finix-S1[8]、DianJin-R1[9]、Fin-R1[10]

Their detailed information is shown as below:

**Table 2 CUFEInse v1.0 Evaluated Model Information**

| Model Name | Parameter Scale | Model Type | Open Source | Reasoning Support | Core Technical Characteristics |
|---|---|---|---|---|---|
| **Gemini-2.5-Pro-0617** | N/A | General-Purpose | No | Yes | Strong general knowledge breadth, outstanding cross-domain reasoning |
| **GPT-4o-1120** | N/A | General-Purpose | No | Yes | Strong multimodal understanding, high general |
| | | | | | generation quality |
| **DeepSeek-R1-0528** | 671B | General-Purpose | Yes | Yes | Optimized for general reasoning, supports long-text processing |
| **GPT-oss-120b** | 120B | General-Purpose | Yes | Yes | Relatively broad financial knowledge coverage among open-source general models |
| **Qwen3-235B-A22B-think** | 235B | General-Purpose | Yes | Yes | Equipped with reasoning mechanism, supports multi-step task decomposition |
| **Qwen3-235B-A22B-instruct** | 235B | General-Purpose | Yes | No | No reasoning mechanism, focuses on instruction following |
| **Qwen3-32B-think** | 32B | General-Purpose | Yes | Yes | Optimized reasoning version for medium-small parameters |
| **Qwen3-32B-instruct** | 32B | General-Purpose | Yes | No | Basic version for medium-small parameters |
| **AntGroup Finix-S1** | N/A | Domain-Specific | No | Yes | Fine-tuned based on insurance business data, focuses on compliance & claims |
| **DianJin-R1** | 32B | Domain-Specific | Yes | Yes | Covers the entire financial domain, with targeted optimization for insurance sub-modules |
| **Fin-R1** | 7B | Domain-Specific | Yes | Yes | Lightweight design, suitable for deployment in small-medium institutions |

### 3.1.2 Evaluation Process Control

To ensure the objectivity of results, a process of "unified input - automatic scoring + manual review" was adopted:

1) **Input Standardization:** The same prompt template was used for all models (e.g., the few-shot mode includes 3 example questions, and the zero-shot mode only includes the question stem) to avoid the impact of prompt deviation on results;

2) **Automatic Scoring:** Multiple-choice questions and judgment questions were automatically scored through answer matching. For short-answer questions and reasoning questions, scoring rules were designed based on "keyword matching + logical integrity" (e.g., actuarial questions require three elements: formula, calculation steps, and results);

3) **Manual Review:** For questions with an automatic scoring error rate exceeding 5% (such as complex clause interpretation questions), two insurance discipline experts conducted independent reviews, and the average score was taken as result.

## 3.2 In-depth Analysis of Evaluation Results by Dimension

### 3.2.1 Insurance Theoretical Knowledge Dimension: Professional Depth Differentiation, Actuarial Ability as a Shortcoming

This dimension evaluates the model's mastery of basic insurance theories and professional knowledge. The core findings are as follows:

1) **Significant Differentiation Between Domain-specific and General Models:** Gemini-2.5-Pro-0617 ranked first with 88.00 points, performing outstandingly in insurance actuarial (86.06 points) — it was the only model with an actuarial sub-item score exceeding 80 points, benefiting from its broad general knowledge and cross-domain reasoning ability; AntGroup Finix-S1 (85.28 points) ranked first in insurance law (93.05 points) and insurance salesperson examination (89.67 points), reflecting the strengthening effect of domain fine-tuning on practical knowledge.

2) **Widespread Weakness in Actuarial Ability:** Most models scored below 40 points in actuarial, such as GPT-4o-1120 (29.83 points), Qwen3-235B-A22B-instruct (32.35 points), and Fin-R1 (15.98 points). The core reason is that actuarial involves complex mathematical formulas (e.g., modified life insurance premium calculation, unearned premium reserve assessment), and the proportion of professional quantitative knowledge in the pre-training corpus of general models is low, with insufficient ability to decompose actuarial logic[16,17].

3) **Significant Improvement from Reasoning Mechanism:** The "reasoning version" of the same basic model outperformed the "non-reasoning version" — for example, Qwen3-32B-think (53.79 points) significantly outperformed Qwen3-32B-instruct (37.01 points) in the actuarial sub-item. This indicates that the reasoning mechanism can help the model decompose multi-step calculation tasks (e.g., step-by-step derivation of "premium = net premium + loading premium") and reduce logical errors.

### 3.2.2 Insurance Industry Understanding Dimension: Scenario Adaptability Determines the Upper Limit of Capability

This dimension focuses on the model's understanding of the entire insurance business chain. The core findings are as follows:

1) **Medical Scenarios as a Shortcoming of General Models:** Non-domain models generally scored below 60 points in the sub-item of disease risk prediction based on physical examination reports, such as DeepSeek-R1-0528 (49.15 points) and Gemini-2.5-Pro-0617 (55.09 points). The reason is that the cross-disciplinary knowledge between medical care and insurance (e.g., "the impact of grade 2 hypertension on critical illness insurance underwriting") requires practical experience, and general models lack such scenario-based training data.

2) **Uneven Coverage of Business Processes:** Non-domain models scored relatively low in insurance liability analysis (average 60.1 points), failing to accurately distinguish between "the liability boundary of critical illness insurance and medical insurance," which reflects the insufficient understanding of insurance business details by general models.

### 3.2.3 Insurance Safety and Compliance Dimension: Copywriting Compliance as a Common Industry Pain Point

This dimension evaluates the model's adherence to regulatory policies and ethical bottom lines. The core findings are as follows:

1) **Copywriting Compliance as the Biggest Risk Point:** The highest score of all models in the "insurance copywriting compliance" sub-item was only 83.18 points (AntGroup), and the lowest was 46.83 points (Fin-R1). Typical problems include: general models tend to generate non-compliant content such as "claim upon diagnosis" (ignoring exclusion clauses) and "minimum interest rate guarantee" (violating insurance product promotion regulations), reflecting that the compliance risks of AIGC in insurance marketing need to be urgently controlled.

2) **Deviation in Value Transmission:** Some models insufficiently transmit the core value of insurance as "risk protection," overemphasizing "profitability" when recommending products while ignoring the "protection function," which is inconsistent with the essential positioning of the insurance industry.

### 3.2.4 Insurance Intelligent Agent Application Dimension: Significant Differentiation in Decision-making and Reasoning Abilities

This dimension evaluates the model's practical service capabilities as an "insurance intelligent agent," with the following key findings:

1) **Inadequate Decision-Making Capabilities of Non-Domain Models:** General-purpose models performed poorly in the underwriting and claim settlement decision sub-item. For example, Qwen3-235B-A22B-instruct (60.49 points) failed to integrate "customer health declaration (hypertension) + product underwriting rules (e.g., underwriting conditions for hypertension in a critical illness insurance)" to provide accurate underwriting recommendations, reflecting its lack of ability to associate business rules with data.

2) **Polarization in Tool Calling Capabilities:** DeepSeek-R1-0528 ranked first in insurance tool parameter extraction (97.02 points), accurately extracting parameters required for the "premium calculator" (e.g., age, sum insured, payment term), but performed moderately in personalized tasks such as product recommendation (74.43 points). Fin-R1, due to parameter scale constraints, had lower accuracy in insurance type comparison than large-parameter models, reflecting the adaptation and balance issue between "parameter scale and scenario complexity."

### 3.2.5 Insurance Logical Rigor Dimension: Numerical Calculation and

## Clause Interpretation as Key Gaps

This dimension assesses the model's logical analysis and accurate interpretation capabilities, with the following key findings:

1) **Domain Models Demonstrate Significant Advantages in Numerical Calculation:** They can accurately calculate "annual returns of participating insurance (including compound interest)" and "surrender cash value (deducting arrears and interest)" with unambiguous clause interpretation (e.g., clearly defining the scope of "war exclusion in exclusion clauses").

2) **Robust Basic Logical Judgment of General-Purpose Models:** GPT-oss-120b performed excellently in insurance question identification (96.16 points), accurately distinguishing between "insurance-related questions (e.g., 'How to reimburse medical insurance?')" and "non-insurance questions (e.g., 'Bank financial product returns')", demonstrating the advantages of general-purpose models in basic logical judgment. DianJin-R1 (90.66 points) also performed well, indicating that the logical rigor of financial domain models can be partially transferred to insurance scenarios.

# 3.3 Comprehensive Ranking and Echelon Classification

Based on the weighted calculation of scores across the 5 dimensions (20% weight for each dimension), the 11 models were divided into three echelons (Table 3):

**Table 3 CUFEInse v1.0 Model Comprehensive Ranking and Dimension Scores (Full Score: 100 Points)**

| **Tier Division** | **Model Name** | **Comp. Score** | **Ins. Theo. Know.** | **Ins. Ind. Under.** | **Ins. Safe. & Comp.** | **Ins. Agent App.** | **Ins. Log. Rigor** |
|---|---|---|---|---|---|---|---|
| **First Tier (>85)** | AntGroup Finix-S1 | 89.51 | 85.28 | 92.47 | 86.78 | 86.86 | 96.15 |
| | Gemini-2.5-Pro-0617 | 85.11 | 88.00 | 84.52 | 78.80 | 82.55 | 91.69 |
| **Second Tier (80-85)** | DeepSeek-R1-0528 | 84.20 | 84.75 | 82.27 | 75.17 | 83.75 | 95.07 |
| | Qwen3-235B-A22B-think | 83.03 | 85.42 | 80.30 | 78.23 | 83.69 | 87.52 |
| | Qwen3-235B-A22B-instruct | 82.66 | 79.07 | 82.75 | 73.93 | 85.52 | 92.04 |
| | DianJin-R1 | 80.35 | 79.88 | 80.47 | 69.82 | 78.25 | 93.30 |
| | Qwen3-32B-think | 80.34 | 80.62 | 82.48 | 65.63 | 82.68 | 90.28 |
| **Third Tier (<80)** | Qwen3-32B-instruct | 79.87 | 76.61 | 82.37 | 70.80 | 79.90 | 89.65 |
| | GPT-4o-1120 | 79.71 | 72.42 | 82.28 | 69.33 | 84.58 | 89.92 |
| | GPT-oss-120b | 79.41 | 75.47 | 78.16 | 70.82 | 79.14 | 93.46 |
| | Fin-R1 | 71.46 | 66.77 | 74.76 | 67.09 | 66.52 | 82.16 |

*Note: Score classification: 90-100 points (Excellent), 80-89 points (Good), 70-79 points (Average), 0-69 points (Needs Improvement); Dimension scores in the table have been standardized.*

# 4 Key Findings and Industry Implications

## 4.1 Core Findings: Capability Boundaries and Technical Directions of Insurance Large Models

1) **Domain Adaptation is the Core Competitiveness of Insurance Large Models:** The evaluation results confirm that the technical route of "general-purpose model + fine-tuning with insurance business data" can effectively improve scenario adaptability. By integrating insurance regulatory policies, actuarial cases, and claim settlement data, domain-specific models make up for the deficiencies of general-purpose models in professional depth and practical capabilities.

2) **Reasoning Mechanism is Key to Breaking Through Bottlenecks in Complex Tasks:** The "reasoning version" of the same basic model (e.g., Qwen3-32B-think) outperforms the "non-reasoning version" (Qwen3-32B-instruct) in the dimensions of insurance theory and logical rigor. This indicates that the reasoning mechanism can help models decompose multi-step tasks (such as actuarial calculations and underwriting decisions) and reduce logical errors in "one-step reasoning," which is an important direction for future technical optimization.

3) **Three Common Bottlenecks Restrict Industry Application:** Current large models face three major bottlenecks in the insurance field: "weak actuarial capabilities, insufficient copywriting compliance, and biased underwriting and claim settlement decisions." Weak actuarial capabilities stem from the lack of quantitative knowledge and calculation logic; insufficient copywriting compliance arises from the disconnection between regulatory policies and marketing scenarios; decision bias results from the lack of ability to associate business rules with user needs, all of which require targeted breakthroughs.

## 4.2 Industry Implications: Implementation Paths of Insurance Large Models from the Evaluation Results

1) **Insurance Company Selection Strategy:** Large insurance companies can independently develop or select domain-specific models for core scenarios such as underwriting and claim settlement, and compliant copywriting generation; small and medium-sized insurance companies can choose open-source models in the second echelon and reduce deployment costs through lightweight fine-tuning (e.g., supplementing local business data).

2) **Technology Iteration Directions for Technology Enterprises:** Model R&D should focus on three directions: "strengthening actuarial capabilities" (e.g., building a dedicated actuarial dataset and optimizing quantitative calculation logic), "compliant copywriting generation" (e.g., embedding a real-time regulatory policy database and designing a compliance verification module), and "optimizing decision reasoning" (e.g., integrating an insurance business rule engine and improving multi-factor association capabilities).

3) **Key Risk Control Areas for Regulatory Authorities:** Attention should be paid to the compliance risks of AIGC in insurance marketing. It is urgent to promote the implementation of "insurance large model compliance evaluation standards," requiring models to pass special tests such as copywriting compliance and value transmission before going online to avoid industry risks caused by non-compliant promotions.

# 5 Optimization Suggestions and Future Outlook for Insurance Large Models

## 5.1 Hierarchical Suggestions for Model Optimization

Based on the evaluation results, the following differentiated optimization suggestions are proposed for different types of models:

1) **Domain-Specific Models:**

- **Deepen Actuarial Professional Depth:** Collaborate with actuarial associations to build an "actuarial case dataset" (covering scenarios such as reserve calculation and risk pricing) and optimize the model's ability to understand and calculate complex mathematical formulas;
- **Expand Coverage of Niche Scenarios:** Supplement data on niche scenarios such as pet insurance, agricultural insurance, and catastrophe insurance to avoid the problem of "sufficient adaptation in mainstream scenarios but insufficient capabilities in niche scenarios."

2) **Non-Domain Models:**

- **Strengthen Scenario-Based Training:** Build a "full insurance business process dataset" (covering insurance application questionnaire analysis, underwriting rule matching, claim settlement document review, etc.) to improve the model's adaptability to practical scenarios;
- **Promote Interdisciplinary Knowledge Integration:** Integrate cross-disciplinary data such as "insurance + medical care" and "insurance + law" (e.g., medical insurance policies, insurance litigation cases) to make up for interdisciplinary knowledge gaps.

3) **Open-Source Models:**

- **Lower the Threshold for Domain Adaptation:** Provide an "insurance domain fine-tuning toolkit," including pre-training weights, special datasets (actuarial, compliance), and fine-tuning tutorials, to help small and medium-sized institutions adapt quickly and promote the ecological construction of open-source insurance large models;

- **Optimize Computational Efficiency:** For lightweight models such as Fin-R1, prioritize supplementing core knowledge such as insurance actuarial and compliant values, and design a "model compression + distillation" scheme to reduce computing power consumption while ensuring core capabilities.

## 5.2 Industry Development Outlook

With the continuous iteration of insurance large model technology, three major trends will emerge in the future:

1) **Integrated Application of "Model + Rule Engine":** The model will be responsible for intent understanding and copywriting generation, while the rule engine will handle underwriting and claim settlement decisions, balancing flexibility and accuracy;

2) **Upgraded Personalized Service Capabilities:** Based on user portraits (e.g., health status, financial goals), the model can provide "one-person-one-policy" insurance solutions, promoting the transformation of insurance services from "standardized" to "personalized";

3) **Deepened Cross-Domain Collaboration:** Cross-domain models of "insurance + medical care + elderly care" will be gradually implemented, realizing integrated services of "health management - risk protection - elderly care planning" and enhancing industry value.

# 6 Conclusion

This paper systematically elaborates on the design scheme and v1.0 evaluation results of the CUFEInse insurance large model evaluation benchmark developed by Central University of Finance and Economics. Through the scientific design of 5 dimensions and 14,430 questions, CUFEInse fills the gap in professional evaluation benchmarks in the insurance field. Its scoring mechanism of "equal weight for dimensions and balance for subcategories" ensures the fairness and interpretability of the evaluation results. The first evaluation results show that domain-specific models are more adaptable to insurance scenarios than general-purpose models, and actuarial capabilities, compliant copywriting generation, and decision reasoning are the core shortcomings of current models.

CUFEInse not only provides a standardized tool for the technical iteration of insurance large models but also offers data support for industry implementation. In the future, with the continuous iteration of the benchmark and the construction of an industry collaborative ecosystem, insurance large models will gradually break through technical bottlenecks and play a greater role in the entire process of product design, precision marketing, intelligent underwriting, and claim settlement services, promoting the high-quality development of the insurance industry towards "intelligence, personalization, and compliance".